%% file: main.tex
\documentclass[10pt,twocolumn,letterpaper]{article}

\usepackage{cvpr}              
\usepackage{graphicx}
\usepackage{multirow}

\definecolor{cvprblue}{rgb}{0.21,0.49,0.74}
\usepackage[pagebackref,breaklinks,colorlinks,allcolors=cvprblue]{hyperref}

\def\paperID{13386} 
\def\confName{CVPR}
\def\confYear{2025}

\title{EchoONE: Segmenting Multiple echocardiography Planes in One Model}

\author{Jiongtong Hu\textsuperscript{1}, 
Wufeng Xue\textsuperscript{1}, 
Jun Cheng\textsuperscript{1}, 
Yingying Liu\textsuperscript{2}, 
Wei Zhuo\textsuperscript{1}\thanks{Corresponding author},
Dong Ni\textsuperscript{1} \\
\textsuperscript{1} Shenzhen University\quad
\textsuperscript{2} Shenzhen People’s Hospital\\
{\tt\small \{weizhuo, xuewf\}@szu.edu.cn}
}

\begin{document}
\maketitle

\input{sec/0_abstract}    
\input{sec/1_intro}
\input{sec/2_related_work}

\input{sec/3_method}

\input{sec/4_experiment}

\input{sec/5_consulsion}
\input{sec/6_acknowledgement}
{
    \small
    \bibliographystyle{ieeenat_fullname}
    \bibliography{main}
}


\end{document}

%% file: sec/0_abstract.tex
\begin{abstract}
In clinical practice of echocardiography examinations, multiple planes containing the heart structures of different view are usually required in screening, diagnosis and treatment of cardiac disease. AI models for echocardiography have to be tailored for each specific plane due to the dramatic structure differences, thus resulting in repetition development and extra complexity. 
Effective solution for such a multi-plane segmentation (MPS) problem is highly demanded for medical images, yet has not been well investigated. 
In this paper, we propose a novel solution, \emph{EchoONE}, for this problem with an SAM-based segmentation architecture, a prior-composable mask learning (PC-Mask) module for semantic-aware dense prompt generation, and a learnable CNN-branch with a simple yet effective local feature fusion and adaption (LFFA) module for SAM adapting. 
We extensively evaluated our method on multiple internal and external echocardiography datasets and achieved consistently state-of-the-art performance for multi-source datasets with different heart planes. This is the first time the MPS problem has been solved in one model for echocardiography data. The code will be available at \href{https://github.com/a2502503/EchoONE} {https://github.com/a2502503/EchoONE}. 
 


\end{abstract}

%% file: sec/1_intro.tex
\section{Introduction}
\label{sec:intro}

Learning-based segmentation methods have been well developed for different medical images to help assist disease screening, diagnosing, and treatment~\cite{TransformerInMedicalSurvey, CNNInMedicalSurvey}. It's usually the case that an organ is scanned from multiple planes of different orientations. For example, in the echocardiography examination, various long- and short-axis planes were required to perform a reliable evaluation of the structure and functionality of the heart~\cite{mitchell2019guidelines}. Segmentation models for echocardiography have to be tailored for each specific plane due to dramatic structure differences, thus resulting in repetition development and extra complexity. A uniform model for multiple plane segmentation (MPS) is highly demanded to easy the development and deployment. However, this problem has not been well investigated.

\begin{figure}[t]
  \centering
   \includegraphics[width=1.0\linewidth]{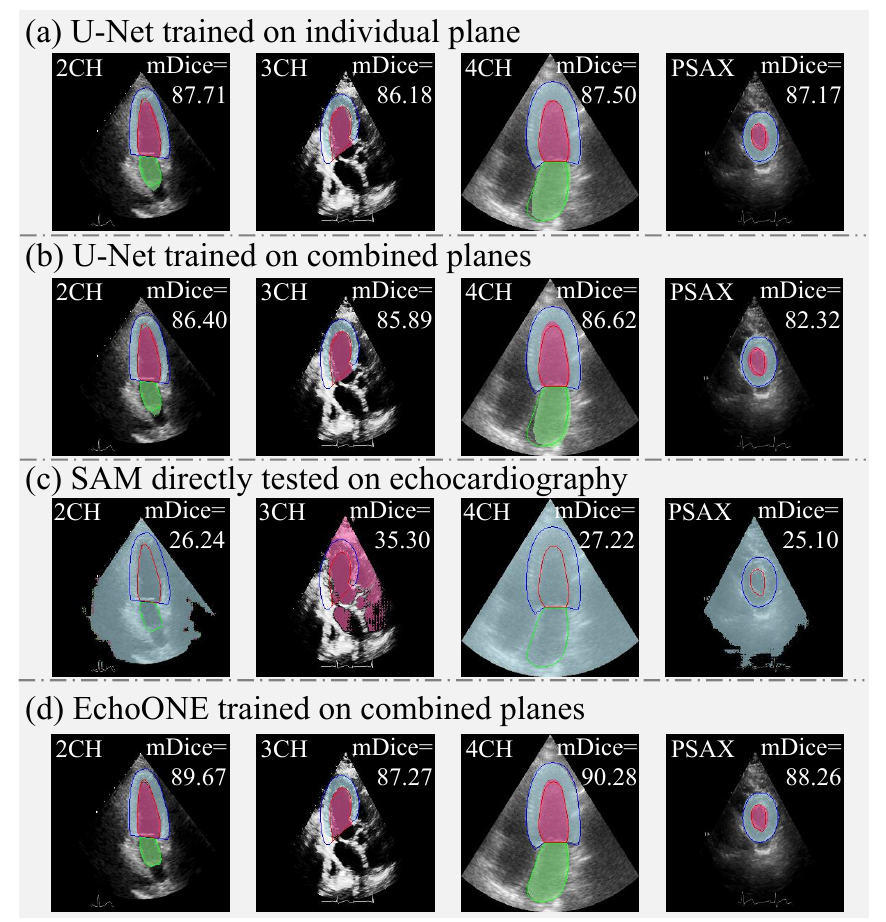}

   \caption{Comparison of existing solutions for the Multi-plane segmentation problem in echocardiographic images. The performance of classic U-Net significantly decreases when trained on multiple planes together (b), compared to training on each plane individually (a). The direct application of SAM on different echocardiographic planes yields poor results (c). EchoONE demonstrates excellent segmentation performance across all planes (d).}
   \label{fig:fig1}
\end{figure}

Existing solutions for the MPS problem include multi-branch architecture in model design and combination of multiple plane data for training. The multi-branch architecture usually leverage some cross-view attention/consistency~\cite{Gvcnn, DRCNN} to help improve the performance. Each plane is still treated separately, and is difficult to generalize to other new planes. When multiple planes are combined to train a uniform segmentation model, performance usually drops obviously when comparing the plane-tailored model. As shown in Row 1 and Row 2 of Figure~\ref{fig:fig1} the results for segmentation of four different heart planes in echocardiography. The model trained with all planes together obtains clearly inferior performance to the plane-specific model. 

Segment Anything Model (SAM) \cite{SAM}, which generalizes well for natural images and may be a potential solution for the MPS problem. However, it fails for medical images due to the large semantic gap. Row 3 of Figure~\ref{fig:fig1} demonstrated the results of SAM when tested with images of the four echocardiographic planes. Without fine-tuning, SAM fails for all these planes. 
Existing SAM-derived methods in medical images focus on adapting/fine-tuning to a specific scenario \cite{Ma-sam, Open-vocabulary-sam} or proposing new prompting generation ways \cite{VRP-SAM, aaai-SAM}. The MPS problem has not been focused in these efforts despite its potential. 

In this paper, we propose a novel solution, \emph{EchoONE}, with a SAM-based segmentation architecture, a prior-composable mask learning (PC-Mask) module for semantic-aware dense prompt generation, and a learnable CNN-branch with a local feature fusion and adaption (LFFA) module for SAM adapting. 

The contribution of this paper can be summarized as:
\begin{enumerate}
\item We present EchoONE, a novel SAM-based model that is capable of accurately segment heart structures from different echocardiographic planes in one model. This is the first uniform model for the multiple plane segmentation problem in medical images. 
\item We propose a prior-composable mask learning (PC-Mask) module for semantic aware dense prompt generation without knowing the plane information of the input, therefore making our EchoONE a uniform model for all planes. 
\item We introduce a local feature fusion and adaption (LFFA) module for feature interaction and fusion with the image encoder and the mask decoder, which helps improve segmentation and accelerate convergence.
\item We design a uniform mask representation for the multi-source echocardiography datasets that using various annotation protocols, enabling the uniform training and validation of our EchoONE model. 

\end{enumerate}

The results of internal and external validation using three public datasets, and four private datasets demonstrate that our EchoONE can achieve consistently state-of-the-art segmentation performance for left ventricle, left atrium, and myocardium from six echocardiographic planes. 





%% file: sec/2_related_work.tex
\section{Related Work}
\label{sec:formatting}

\subsection{Multi-plane segmentation}
Multi-plane analysis of medical images enables a comprehensive evaluation of the organ of interest in clinical practice, therefore raising the requirement for multi-plane segmentation (MPS) of medical images.
It is noteworthy that the challenge of multi-plane analysis transcends medical imaging domains. \cite{SPIn-NeRF} integrates a semantic neural radiance field in conjunction with a video instance segmentation model to segment the appearance of the same scene from multiple view directions. \cite{radar_rebuttal} uses three encoders with a shared latent space for multi-view radar segmentation.
Existing solutions in the domain of medical image segmentation can be attributed into two categories: multi-branch architecture with each branch dealing with one plane, and a uniform model that is trained using combination of multi-plane image-mask pairs. While the latter often leads to performance drop, efforts have been devoted to the former one. 
\cite{ding2021mvfusfra} designed a network with three segmentation branches for each of the sagittal, coronal, and axial planes for brain tumor segmentation. Then fused 3D results of the three planes were then employed to improve the segmentation of each plane in turn. \cite{zhong2025cross} presents multi-plane representation learning for volume medical data by exploring plane discrepancy and dependency within three parallel branches for sagittal, coronal, and axial views, resulting in improved 3D segmentation. TransFusion \cite{liu2022transfusion} introduced multiscale cross-view context modeling and dependency mining between two parallel branches for the segmentation of short and long axis cardiac MRI images. 

Similarly, in cardiac structure segmentation, there are studies that combine information from multiple planes \cite{JBHI-MMs2}. For instance, \cite{MMs2workshop1} aligns segmentation information from the long-axis view (LA) with the short-axis view (SA), while \cite{MMs2workshop2} shares low-level features between LA and SA to achieve right ventricle (RV) segmentation in cardiac MRI. 

To summarize, these methods follow a divide-and-connection way to deal with the multiple planes, i.e., treating different planes separately first and then try to improve the segmentation through cross-view dependency/attention modeling, feature sharing and combination. They are intrinsically not a uniform model for MPS, and the generalization to other new planes cannot be guaranteed. In this paper, we treat all planes with one uniform model, and validate it with echocardiographic planes, which is even more challenging due to its non-standard scanning procedure. 

\begin{figure*}[http]
    \centering
    \includegraphics[width=0.91\textwidth]{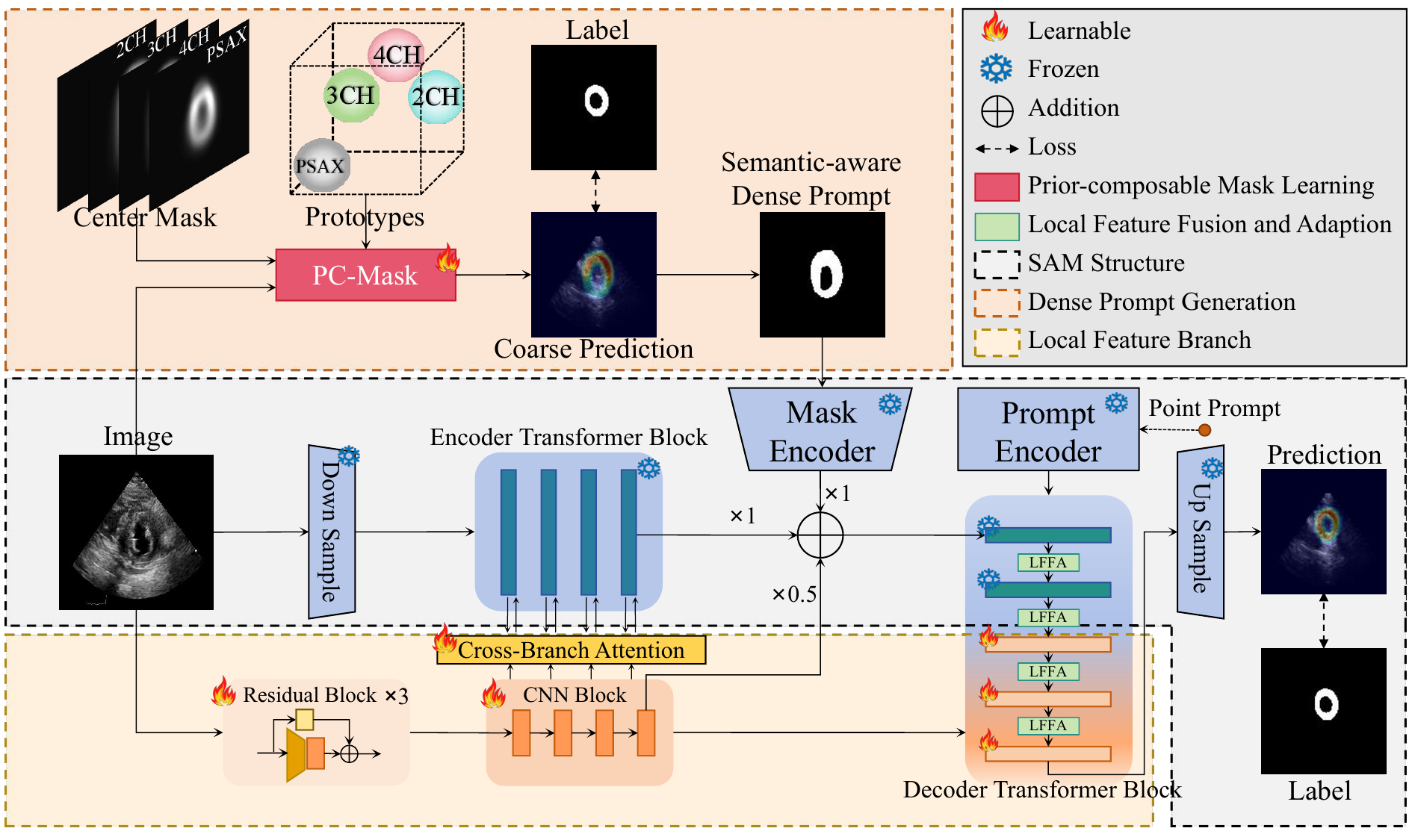}  
    \caption{Overview of our EchoONE, which contains a SAM-based segmentation architecture (middle), a prior-composable mask learning (PC-Mask) module (top) for dense prompt generation, and a learnable CNN-based local feature branch (bottom) for SAM encoder tuning and decoder adapting.
    }
    \label{fig:overview}  
\end{figure*}

\subsection{SAM in medical image tasks}
SAM is a robust zero-shot segmentation foundation model that has demonstrated exceptional performance on various natural image tasks \cite{zhang2023sam3d, wang2024samrs, osco2023segment}. However, directly applying SAM to medical imaging often results in a significant loss of generalization capability \cite{huang2024segment, mazurowski2023segment, chen2023ability}. This is largely due to the fact that, in contrast to natural images, medical images typically contain low contrast structures, high level noises, indistinct tissue boundaries, thus leading to inherent disparities in data distribution. 
Currently, many studies focus on fine-tuning SAM to restore its robust segmentation capability in medical imaging tasks. MedSAM \cite{MedSAM} constructed a large-scale medical image dataset and utilized bounding box prompts to guide SAM. AutoSAM \cite{AutoSAM} replaces the interactive input prompts with a learnable encoder, enabling a fully automated procedure. SAMed \cite{SAMed} applies a low-rank-based (LoRA) fine-tuning strategy to the image encoder and uses the default prompt embedding to achieve prompt-free results. SAMMed-2D \cite{SAMMed-2D} and MSA \cite{MSA} both employ Adapter blocks to fine-tune SAM. SAMMed-2D adds an Adapter Layer to each transformer block of the image encoder, adjusting the feature across both channel and spatial dimensions. MSA integrates prompt information into the adaptation process. SAM-LST \cite{SAM-LST} is inspired by another fine-tuning method, Ladder Side-Tuning (LST), and introduces an additional CNN encoder, providing a new approach for combining CNN with SAM. SAMUS \cite{SAMUS} employs both LST and Adapter Tuning methods, using position and feature adapters to transition SAM to medical domains. 

In summary, these SAM-based methods facilitate the transition of SAM from natural images to medical images, demonstrating strong feature extraction and generalizability. However, none have been designed for the MPS problem, especially for the multiple echocardiography planes, where significant structural differences exist between planes and the heavy noise and low contrast may deteriorate the prompting and the mask decoder in SAM.



\subsection{Prompt engineering in SAM}
The configuration of the prompts significantly impacts the effectiveness of SAM models \cite{huang2024segment}, and has become the focus of recent efforts. SAMAug \cite{SAMAug} generates enhanced point prompts that provide more comprehensive user intent information. For remote sensing instance segmentation, RSPrompter \cite{RSPrompter} introduces a class-based prompt generator, enabling the output mask to align with the specific global semantics. PA-SAM \cite{PA-SAM} extracts information from the image for a prompt adapter, generating additional sparse and dense prompts to optimize the decoder. In the medical imaging domain, MSA \cite{MSA} constructs an adapter module focused on prompt integration, while ProMISe \cite{ProMISe} experiments with combining point coordinates and image embeddings to refine prompt information. SurgicalSAM \cite{SurgicalSAM} integrates surgical instrument-specific information with SAM's pre-trained knowledge via a learnable prototype-based class prompt module, achieving efficient and lightweight tuning. 

These methods add information from the user or the input images to implicitly make a enriched prompt, thus improving the final performance. We explicitly employ prior semantic knowledge of different planes to sense the structure of the input image and generate the mask prompt. 

%% file: sec/3_method.tex

\section{Method}
\label{sec:3_method}


\subsection{Overview}
Figure \hyperref[fig:overview]{\textcolor{cvprblue}{2}} shows the overview architecture of EchoONE. EchoONE mainly consists of three components, a SAM-based segmentation architecture, a prior-composable mask learning (PC-Mask) module for dense prompt generation, and a CNN-based local feature branch for SAM tuning and adapting.

The overall network architecture is based on the original SAM, including transformer-based image encoder ($E_{I}$) and mask decoder ($D_M$), a sparse prompt encoder ($E_{SP}$), a mask encoder for dense prompt ($E_M$). In addition to this, we introduce a local feature fusion and adaption (LFFA) module in the LST branch to adapt SAM to our task, and propose a clustering-based PC-Mask module for generation of semantic-aware dense prompt. Details of PC-Mask and LFFA are given below.



\begin{figure}[t]
    \centering
    \includegraphics[width=1.0\linewidth]{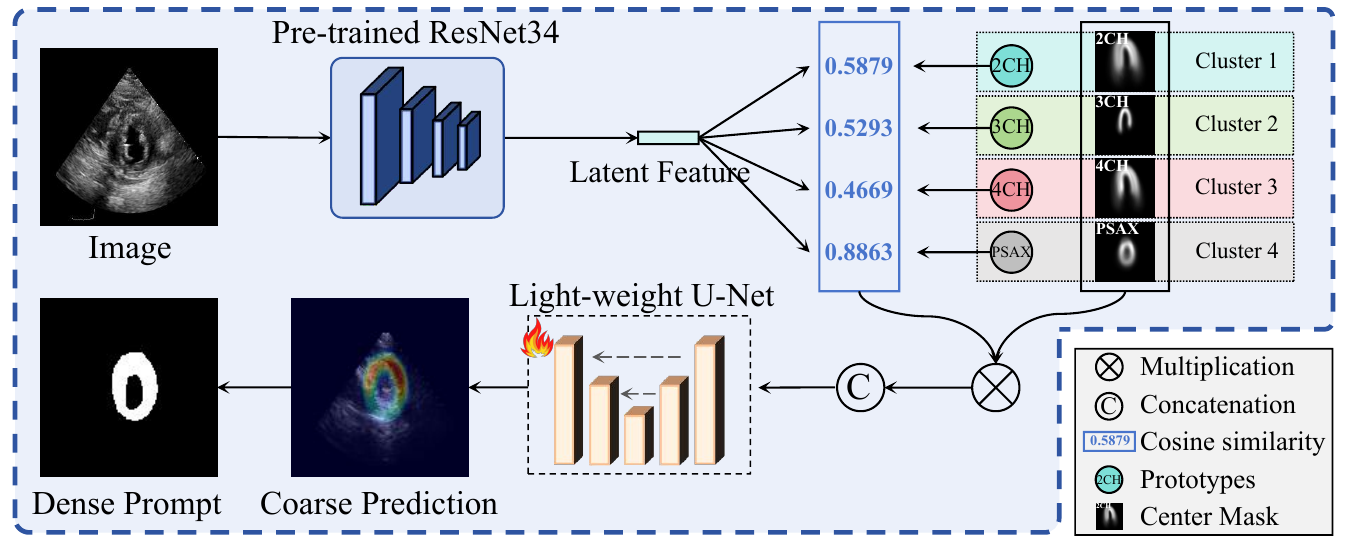}  
    \caption{Details of our prior-composable mask learning (PC-Mask) module, which leverages prior structural information to automatically generate high-quality semantic-aware dense prompts for images without knowing the plane information.}
    \label{fig:PC-Mask}  
\end{figure}
\subsection{Prior-composable mask learning for dense prompt generation}
Dense mask prompt provides richer information to SAM, therefore results in better segmentation than point or box prompts. We propose the PC-Mask module to automatically generate high quality mask prompts with prior structure information embedded for images of unknown plane.


To deal with the diversity of semantic structures in multiple planes, we first group images from different planes into $K$ clusters $C_i, i\in\{1,...,K\}$ in a latent feature space ($E_{Lat}$). The center of each cluster is used as a prototype $u_i, i\in\{1,...,K\}$ for the group in the latent space. Similarly, a center mask $m_i, i\in\{1,...,K\}$ can be obtained by averaging the masks of those images that are assigned to cluster $i$. In our implementation, the latent space is obtained by a pre-trained ResNet34 \cite{ResNet} for plane classification. 

With these mask centers as the structure prior, we aim to customize them for each new image so that the resulted mask is aware of the structure in that image without knowing its plane information. 
Specifically, for an input image $I_j$, its location in the latent space can be represented by its similarity (or distance) to these prototypes $u_i$. In our work, the cosine similarity is used:  
\begin{align}
    w(i,j) &= cossim(E_{Lat}(I_j), u_i) \notag\\
             &= \frac{<E_{Lat}(I_j), u_i>}{\|E_{Lat}(I_j)\|\|u_i\|}    
\end{align}
A latent vector representing the image's location in the latent space can be obtained: $[w(1,j),...w(K,j)]$. 
Then these prior centers are customized into a multichannel prior embedding ($PE_j$) by using the similarity vector as weights:
\begin{equation}
    PE_j = \mathbf{concat}([w(i,j)\times m_i]), i=1,...K
\end{equation}
With the customized prior embedding $PE_j$ as input, we design a light-weight U-Net to learn the mask of the input image using a combination of dice loss and binary cross-entropy Loss $L_{PCM}$. The output $PCM_j$ is used as our dense prompt for SAM. 
\begin{equation}
    PCM_j = UNet_{\theta}(PE_j)
\end{equation}
where $\theta$ is the network weights to be optimized. In this way, diverse semantic-aware masks can be composed and learned.

It's worth to note that our PC-Mask is not derived directly from the input image, therefore the semantics differs from traditional segmentation network and could be an effective guidance for SAM. Besides, no plane information is used during the generation, which makes our model capable of dealing with the MPS problem in one uniform model. 





\subsection{CNN branch for tuning and adaption}

\begin{figure}[t]
    \centering
    \includegraphics[width=0.8\linewidth]{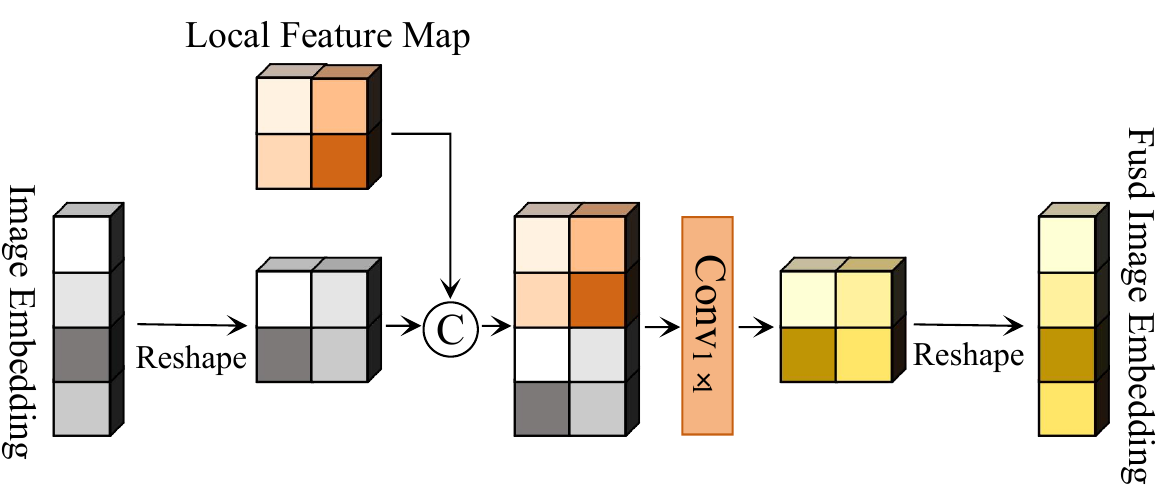}  
    \caption{Local feature fusion and adaption (LFFA). The output of each CNN layer (local feature) and image embeddings from a previous transformer is first concatenated and then mixed with a convolution layer to obtain the fused image embedding.}
    \label{fig:fig2}  
\end{figure}

To fully leverage the capability of SAM without resourcing-intensive retraining, an auxiliary branch for tuning and adapting SAM to a new scenario is necessary. 
Inspired by previous SAM-based work~\cite{SAMUS,SAM-LST}, we design a learnable CNN branch as shown in the bottom of Figure~\ref{fig:fig2} for this. 
Our CNN branch is constituted of three parts: the Residual block for local feature extraction, the CNN block with cross-branch attention for tuning of the image encoder, and the transform blocks to fusion the local feature to adapt the mask decoder to our task.


In addition to the two original transformer blocks in the mask decoder, we add three more learnable transformer blocks in order to accommodate the integration of local features. The local features from the four CNN layers in the CNN blocks are fused connected directly to the last four Transformer blocks via skip connection, respectively. For features in each CNN layer and transformer block, they are fused by a specific fusion and adaption module LFFA, as shown in Figure~\ref{fig:fig2}. Denote $f_{CNN,l}$ as the CNN feature in the \emph{l}th layer of the CNN block, $f_{DM-K,l}$ and $f_{DM-Q,l}$ the outputs of the \emph{l}th transformer block in the mask decoder, and $f_{F,l}$ the fused feature, the procedure of LFFA can be described as:
\begin{equation}
    f_{F,l} = \mathbf{conv}_{1\times1} (\mathbf{concat}(f_{CNN, l}, (f_{DM-K, l})))
\end{equation}
Operation \emph{Reshape} is used when necessary during the fusion. Then $f_{F,l}$ is used as the input for the next transformer block. 
\begin{equation}
    [f_{DM-K,l+1}, f_{DM-Q,l+1}]=\mathbf{DTrans}(f_{F,l},f_{DM-Q,l})
\end{equation}
where $\mathbf{DTrans}$ is the transformer block in the mask decoder of SAM. Here the other two inputs of the transformer block, i.e., the position embedding and the point prompt, are ignored for simplicity. We have observed in our experiments that such an LFFA design not only integrates local feature better, but also accelerates the model's convergence.

\begin{figure}[t]
    \centering
    \includegraphics[width=1.0\linewidth]{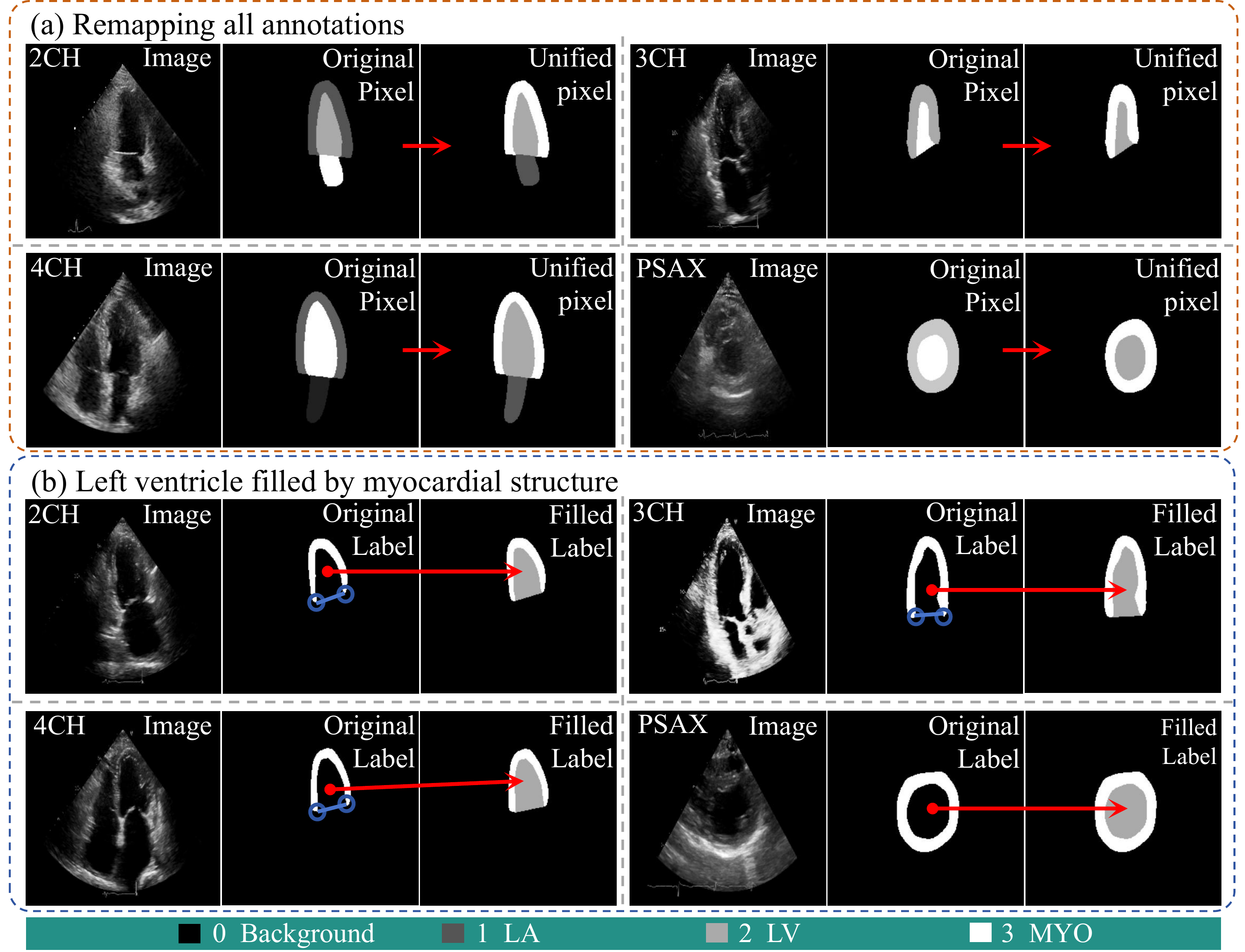}  
    \caption{(a) Remapping annotations to unified mask representation. (b) Preprocessing to generate LV cavity from the annotation of MYO in multiple planes. Landmarks are required to be detected first for planes of 2CH, 3CH and 4CH.}
    \label{fig:fill_mask}  
\end{figure}

\subsection{Overall objective}
To supervised the final segmentation from the mask decoder of SAM, combination of the dice loss \cite{diceloss} and the binary cross-entropy loss with weights of 0.8 and 0.2 is used as the objective function $L_{SEG}$. Plus with the PC-Mask learning objective $L_{PCM}$, the overall objective is:
\begin{equation}
  \mathcal{L}_{EchoONE} = \mathcal{L}_{SEG}+\lambda \mathcal{L}_{PCM}
  \label{eq:overall-loss}
\end{equation}
The tradeoff factor $\lambda$ is set to 0.5, allowing the model to primarily focus on supervising the final segmentation result while effectively encouraging PC-Mask learning.


%% file: sec/4_experiment.tex
\begin{table}[t]
  \centering
  \resizebox{1.0\linewidth}{!}
  {
  \begin{tabular}{ccccc}
    \toprule
    Center & 2CH & 3CH & 4CH & PSAX \\
    \midrule
    CAMUS & \multicolumn{1}{l}{$9268^{(1,2,3)}$} & - & \multicolumn{1}{l}{$9964^{(1,2,3)}$} & - \\
    HMC\_QU & - & - & \multicolumn{1}{l}{$2349^{(3)}$} & - \\
    EchoNet-Dynamic & - & - & \multicolumn{1}{l}{$2552^{(2)}$} & - \\
    \hline
    A & \multicolumn{1}{l}{$3643^{(2,3)}$} & \multicolumn{1}{l}{$3638^{(2,3)}$} & \multicolumn{1}{l}{$3614^{(2,3)}$} & - \\ 
    B & \multicolumn{1}{l}{$1419^{(1,2,3)}$} & \multicolumn{1}{l}{$401^{(2,3)}$} & \multicolumn{1}{l}{$1411^{(1,2,3)}$} & \multicolumn{1}{l}{$930^{(2,3)}$} \\
    C & \multicolumn{1}{l}{$590^{(2)}$} & - & \multicolumn{1}{l}{$1269^{(2)}$} & - \\
    D & \multicolumn{1}{l}{$1774^{(2,3)}$} & \multicolumn{1}{l}{$210^{(2,3)}$} & \multicolumn{1}{l}{$1758^{(2,3)}$} & \multicolumn{1}{l}{$1387^{(2,3)}$}\\
    \bottomrule
  \end{tabular}
  }
  \caption{Summary of the plane and annotation information for three public datasets and four private datasets. The abbreviations A-D represent different datasets. The upper script (1,2,3) are short for annotation of the left atrium, left ventricle and myocardium.}
  \label{tab:dataset}
\end{table}

\section{Experiments and analysis}

\begin{table*}[t]
  \centering
  \resizebox{1.0\textwidth}{!}
  {
  \begin{tabular}{ccccccccccccc}
    \toprule
    \multirow{2}{*}{Method} & \multicolumn{3}{c}{2CH} & \multicolumn{3}{c}{3CH} & \multicolumn{3}{c}{4CH} & \multicolumn{3}{c}{PSAX} \\ \cline{2-13} 
     &mDice&mIoU& HD95 & mDice & mIoU & HD95 & mDice & mIoU & HD95 & mDice & mIoU & HD95\\     
    \midrule
    U-Net\cite{U-Net}      &86.40& 77.05 & 7.53 & 85.89 & 76.02 & 4.84 & 86.62 & 77.96 & 6.79 & 82.32 & 71.64 & 6.65\\
    DeepLabV3+\cite{DeepLabV3Plus} &87.73& 78.92 & 6.82 & 86.54 & 77.01 & 4.76 & 88.35 & 80.13 & 6.18 & 84.40 & 74.16 & 5.84\\
    SwinUNet\cite{SwinUNet}   &86.13& 76.86 & 7.76 & 83.68 & 73.19 & 5.53 & 84.69 & 75.67 & 7.47 & 79.94 & 68.04 & 7.69\\
    H2Former\cite{H2Former}   &88.01& 79.50 & 6.66 & 86.45 & 77.05 & 4.58 & 88.26 & 80.16 & 6.00 & 85.27 & 75.21 & 6.07\\
    \hline
    SAM\cite{SAM}        &26.24& 16.01 & 140.01 & 35.30 & 24.04 & 55.30 & 27.22 & 16.75 & 134.79 & 25.10 & 15.53 & 86.53 \\
    MedSAM\cite{MedSAM} & 81.76 & 70.46 & 11.71 & 82.57 & 71.22 & 7.03 & 84.14 & 73.75 & 9.56 & 76.99 & 63.98 & 9.01\\
    SAMed\cite{SAMed} &81.95& 70.83 & 9.61 & 79.74 & 67.24 & 7.18 & 82.49 & 71.64 & 8.66 & 65.10 & 51.15 & 11.33\\
    SAM-Med2D\cite{SAMMed-2D} &81.59& 72.90 & 10.07 & 74.33 & 65.88 & 8.47 & 82.22 & 73.79 & 8.44 & 73.40 & 62.75 & 9.8\\
    MSA\cite{MSA} &85.88& 76.00 & 9.04 & 80.50 & 68.18 & 6.59 & 85.41 & 75.69 & 8.30 & 76.77 & 63.65 & 9.54\\
    SAMUS\cite{SAMUS} &87.51& 79.38 & 7.57 & 85.54 & 76.17 & 4.92 & 88.37 & 80.35 & 6.54 & 85.92 & 76.52 & 5.57\\
    \hline
    EchoONE &\textbf{89.67}& \textbf{81.87} & \textbf{6.52} & \textbf{87.27} & \textbf{78.01} & \textbf{4.54} & \textbf{90.28} & \textbf{82.67} & \textbf{5.61} & \textbf{88.26} & \textbf{79.68} & \textbf{4.67}\\
    \bottomrule
  \end{tabular}
  }
  \caption{Plane-wise internal evaluation of segmentation performance for EchoONE and the competitors. EchoONE demonstrates its advantages in terms of accuracy for each plane and robustness across multiple planes.}
  \label{tab:all_result}
\end{table*}

\subsection{Datasets and evaluation metrics}
To develop and evaluate our method, we collected multiple echocardiographic datasets with different source and annotation protocols. Five datasets including one public dataset CAMUS \cite{CAMUS} and four private datasets from four medical centers of different cities (denotes as A, B, C, D below) are employed for model training and internal evaluation. Two additional datasets are further used for external evaluation: HMC\_QU \cite{HMC_QU} and EchoNet-Dynamic \cite{EchoNet-Dynamic}. 


\textbf{Dataset details:}
Details of the plane information and annotation can be seen in Table~\ref{tab:dataset}. There are a total of six echocardiographic planes, i.e., apical 2-chamber (2CH), apical 3-chamber (3CH), apical 4-chamber (4CH), parasternal short axis basal level (PSAX-B), parasternal short axis mid-ventricle level (PSAX-M), parasternal short axis apical level (PSAX-A). Given the fact that the three PSAX planes share similarities, we treat them as one PSAX plane in this work. In the table, the image number of each plane in the dataset and how each plane is annotated are demonstrated. Taking CAMUS as an example, two planes of 2CH and 4CH are included. For each image, the left ventricle (LV), myocardium (MYO) and the left atrium (LA) are annotated. 

\textbf{Unified mask representation for inconsistent annotation protocols:} Since our datasets have different sources, contain different planes, the annotation information differs a lot across datasets. It's necessary to first unify their annotation representation so that a uniform model can be developed based on them. In this work, we remap all annotations according to the following protocols: 0 for background, 1 for LV, 2 for LV cavity, and 3 for MYO. Since all planes show the heart structure from a specific orientation, such as protocol is applicable to all planes, and to all datasets as well. Also see Figure~\ref{fig:fill_mask} (a) for examples of this annotation remapping. Besides, for those image with only MYO annotated, we generate the masks for LV cavity by detecting the landmarks and filling the cavity (as shown in Figure~\ref{fig:fill_mask} (b)).  




\begin{figure}[t]
    \centering
    \includegraphics[width=0.95\linewidth]{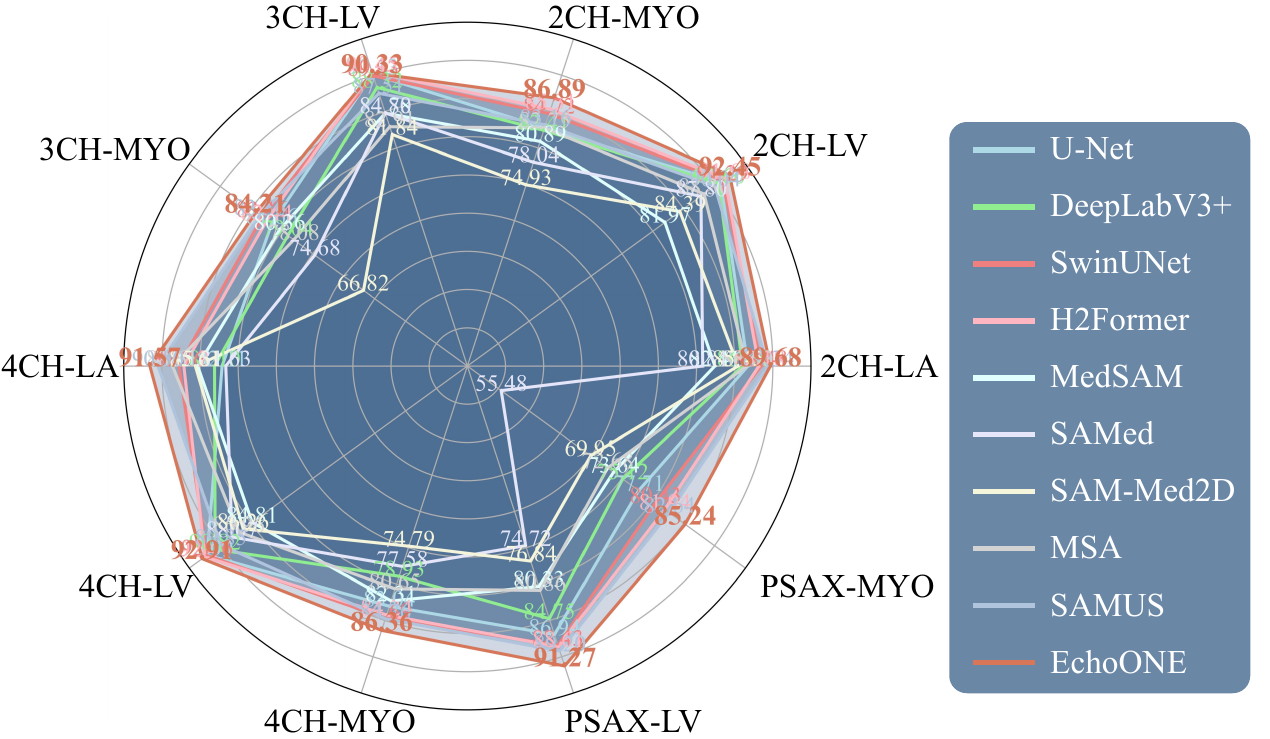}  
    \caption{Structure-wise performance of EchoONE and its competitors. The mean value Dice score of the same structure within one plane is computed across the test data of all internal datasets.}
    \label{fig:latar}  
\end{figure}

\textbf{Training and evaluation:} 
The datasets are used as follows during training and evaluation. Datasets of CAMUS, A, B, C, and D are used as internal data, where images of 80\% subjects for each are randomly selected for model training, 10\% for validation during training, and the rest 10\% are used as internal evaluation. 
Datasets HMC\_QU and EchoNet-Dynamic are only used for external evaluation to test the generalization of our model. 

\textbf{Evaluation metrics:} 
We utilize widely used metrics such as Dice coefficient (mDice) and mean Intersection over Union (mIoU) for segmentation evaluation, along with the Hausdorff Distance-95 \% (HD95).Here the prefix \emph{m} can denote the mean value for each structures, each plane or each dataset.

\subsection{Implementation details}

Before training, the parts inherited from SAM were initialized with the pre-trained weights of SAM on the SA-1B dataset using a ViT-B backbone, while the remaining parameters were initialized randomly. During the training process, we employed a combination of random rotation, scaling, contrast adjustment, and gamma enhancement for data augmentation, each with a probability of 0.5. All images  were resized to 256×256. We trained our datasets for 100 epochs. The Adam optimizer is used, with $\beta_{1}$, $\beta_{2}$ set to 0.9 and 0.999, and an initial learning rate of $1e-4$. All our implementation is in PyTorch and we train our models on single NVIDIA 48G A6000 GPU.

\begin{figure}[t]
    \centering
    \includegraphics[width=0.95\linewidth]{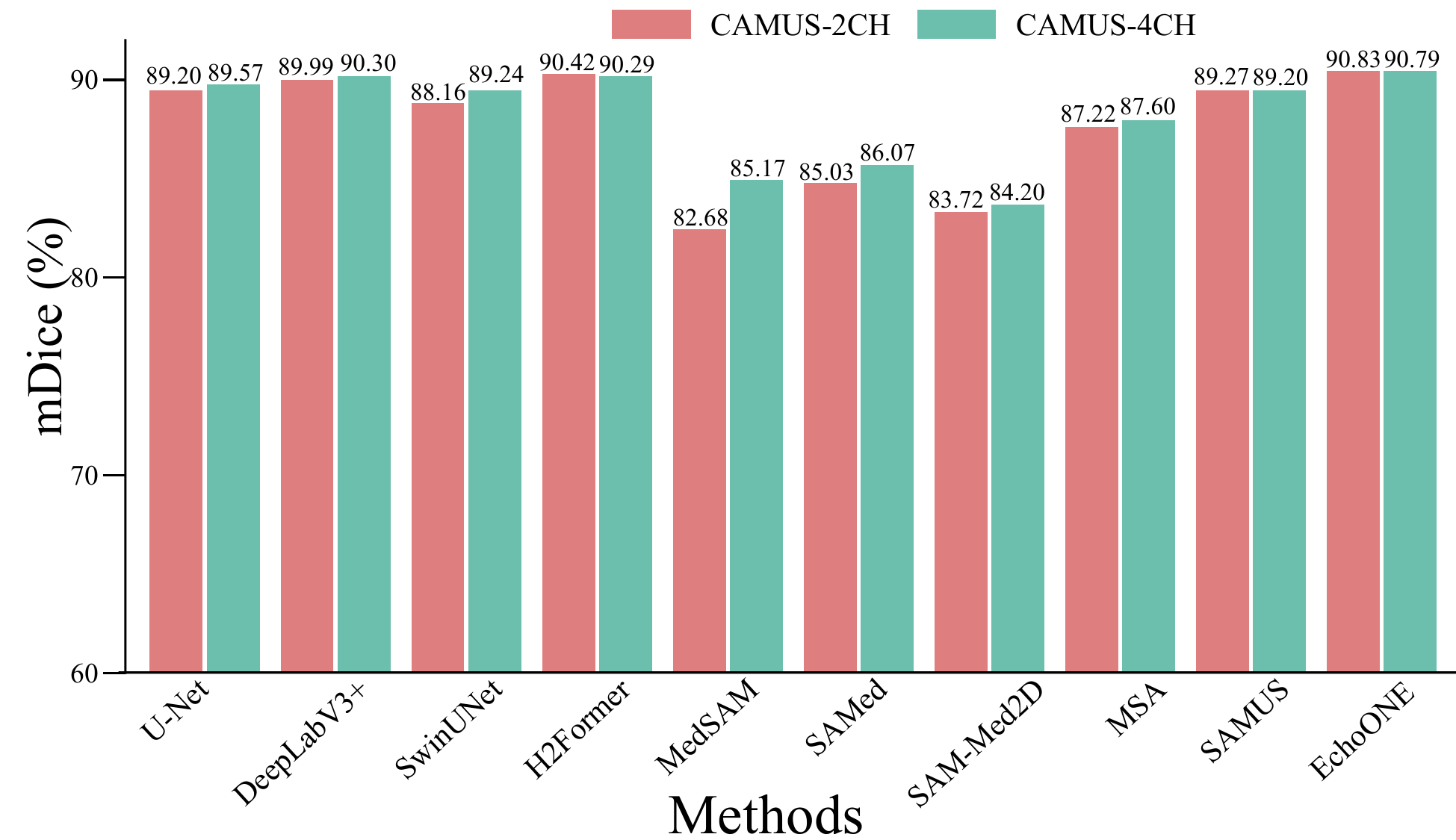}  
    \caption{Performance comparison of EchoONE and its competitors for the internal dataset CAMUS.}
    \label{fig:camus-2ch-4ch}  
\end{figure}

\subsection{Results}

\begin{figure*}[t]
    \centering
    \includegraphics[width=0.95\textwidth]{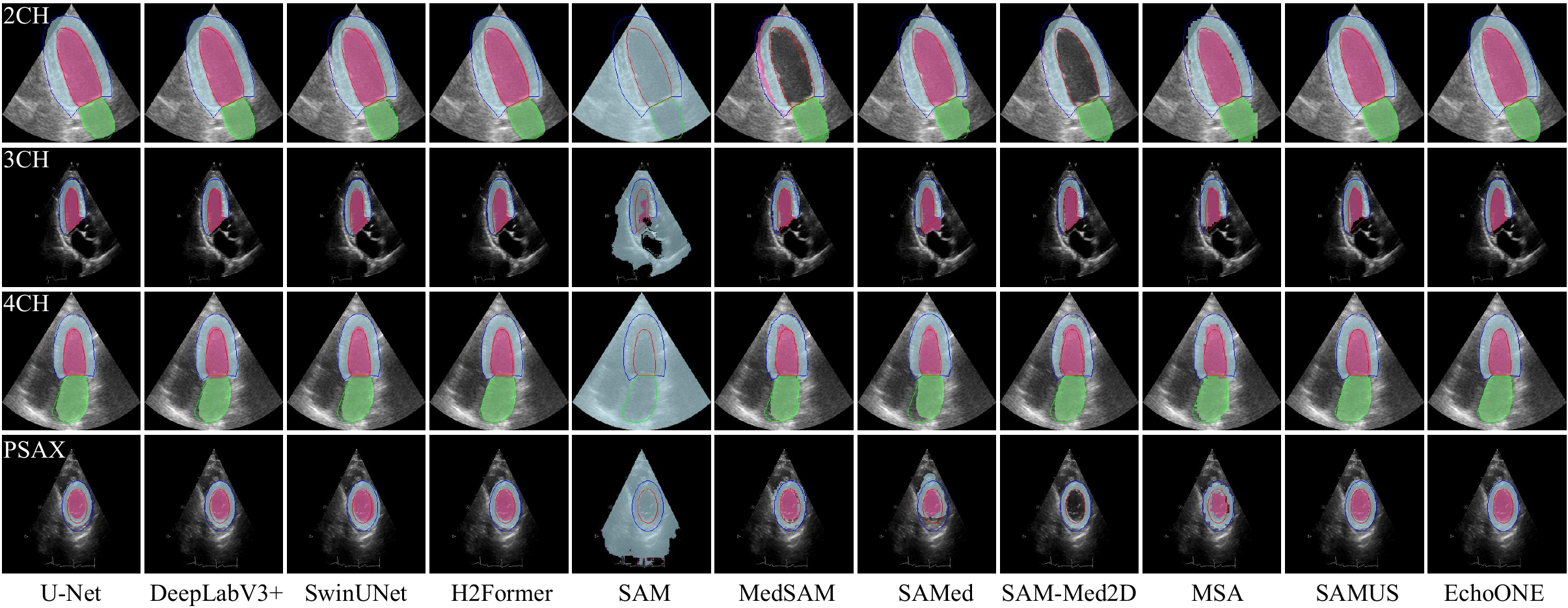}  
    \caption{Visualization of the segmentation results by EchoONE and existing methods. The green, blue, and red lines correspond to the ground truth contours of LA, LV, and MYO, respectively, while the light-colored regions denote the predicted masks.}
    \label{fig:all_model_show}  
\end{figure*}

We conducted extensive experiments using internal and external datasets to demonstrate the effectiveness of our method for the MPS problem. Existing methods used for comparison include four traditional models: CNN-based U-Net \cite{U-Net} and DeepLabV3+ \cite{DeepLabV3Plus}, transformer-based SwinUNet \cite{SwinUNet}, and the CNN-transformer hybrid model H2Former \cite{H2Former}, as while as six SAM-based methods: the original SAM \cite{SAM}, MedSAM \cite{MedSAM}, SAMed \cite{SAMed}, SAM-Med2D \cite{SAMMed-2D}, MSA \cite{MSA}, and SAMUS \cite{SAMUS}. Except SAM, which is applied directly to the test data, the rest of the methods are trained using the same datasets as those used to train EchoONE.

\subsubsection{Internal evaluation}

\textbf{Robustness to multiple planes:}
We first show the results for internal evaluation in Table~\ref{tab:all_result} in an plane-wise manner to better evaluate the robustness of these methods for MPS problem. For each plane, the results from the five internal datasets are averaged. 
Among the four traditional models, U-Net, SwinUNet, and H2Former generally perform better on our dataset compared to most of the SAM-based models, with the exception of SAMUS, which is specifically designed for ultrasound image segmentation. The rest SAM-based methods, although retrained using the same data, only achieve reasonable good performance, and show clear lower performance than EchoONE for all the planes. It's worth to note that EchoONE performs significantly better than all competitors for the PSAX plane, which may be attributed to the uniform design of the semantic-aware PC-Mask for dense prompt.    


\textbf{Robustness across different structures:}
Figure \ref{fig:latar} presents a more detailed Dice (\%) score for each cardiac structure (LV, LA, MYO) of different planes. It can be drawn that our method consistently achieves a higher Dice score than all competitors for all these cardiac structures.

\textbf{Robustness across datasets:}
We also display the dataset-wise results in Table~\ref{tab:ABCD} for A, B, C, D, and Figure~\ref{fig:camus-2ch-4ch} for CAMUS. Average is taken across planes of each dataset. The proposed EchoONE outperforms all competitors in all five internal datasets.  

\subsubsection{External evaluation}
Results for external evaluation are shown in Table~\ref{tab:external_dataset_perform}. Without training with the two datasets, our method performs the best for both of them, revealing its great generalization. Even for HMC\_QU, where low quality images exist and annotations are noisy, our method still delivers a Dice score of 73.94\%. These observations demonstrate the great potential of EchoONE for images of real clinical practice.

\begin{table*}[h]
  \centering
  \resizebox{1.0\textwidth}{!}
  {
  \begin{tabular}{ccccccccccccc}
    \toprule
    \multirow{2}{*}{Method} & \multicolumn{3}{c}{A} & \multicolumn{3}{c}{B} & \multicolumn{3}{c}{C} & \multicolumn{3}{c}{D} \\ \cline{2-13} 
     &mDice&mIoU& HD95 & mDice & mIoU & HD95 & mDice & mIoU & HD95 & mDice & mIoU & HD95\\     
    \midrule
    U-Net\cite{U-Net}      &86.75& 77.08 & 4.46 & 77.35 & 66.95 & 7.86 & 79.21 & 68.20 & 6.82 & 87.51 & 78.63 & 5.50\\
    DeepLabV3+\cite{DeepLabV3Plus} &87.26& 77.98 & 4.69 & 83.04 & 73.33 & 6.29 & 81.83 & 70.99 & 6.54 & 88.31 & 79.97 & 4.95\\
    SwinUNet\cite{SwinUNet}   &86.39& 76.59 & 4.85 & 77.30 & 67.29 & 8.16 & 80.24 & 68.89 & 7.04 & 87.18 & 78.34 & 5.68\\
    H2Former\cite{H2Former}   &87.28& 77.86 & 4.57 & 83.85 & 74.63 & 6.02 & 83.18 & 72.67 & 6.24 & 89.18 & 81.37 & 4.47\\
    \hline
    SAM\cite{SAM}        &37.01& 25.11 & 50.09 & 23.36 & 14.11 & 104.53 & 31.53 & 20.79 & 68.71 & 29.21 & 18.22 & 101.73 \\
    MedSAM\cite{MedSAM} &81.23& 69.98 & 6.51 & 76.57 & 63.45 & 6.74 & 81.00 & 69.42 & 6.55 & 84.14 & 74.58 & 5.83\\
    SAMed\cite{SAMed} &81.54& 69.48 & 6.27 & 69.19 & 56.60 & 10.10 & 69.04 & 55.29 & 9.16 & 80.82 & 69.34 & 7.81\\
    SAM-Med2D\cite{SAMMed-2D} &78.12& 67.97 & 7.07 & 76.02 & 67.23 & 8.10 & 71.55 & 59.75 & 9.57 & 86.72 & 78.02 & 5.14\\
    MSA\cite{MSA} &81.04& 69.31 & 6.61 & 81.93 & 71.02 & 7.57 & 78.74 & 66.39 & 8.30 & 86.48 & 76.95 & 5.52\\
    SAMUS\cite{SAMUS} &85.82& 76.69 & 4.95 & 86.49 & 77.72 & 5.46 & 81.91 & 71.58 & \textbf{5.12} & 88.18 & 80.29 & 4.85\\
    \hline
    EchoONE &\textbf{88.39}& \textbf{79.64} & \textbf{4.27} & \textbf{88.90} & \textbf{80.76} & \textbf{4.23} & \textbf{85.54} & \textbf{75.82} & 5.45 & \textbf{90.63} & \textbf{83.33} & \textbf{3.89}\\
    \bottomrule
  \end{tabular}
  }
    \caption{Dataset-wise results of EchoONE and other models on the evaluation set of the four private internal datasets.}
  
  \label{tab:ABCD}
\end{table*}

\begin{table}[t]
  \centering
  \begin{tabular}{ccccc}
    \toprule
    \multirow{2}{*}{Method} & \multicolumn{2}{c}{HMC\_QU} & \multicolumn{2}{c}{EchoNet-Dynamic} \\ \cline{2-5} 
    & mDice & mIoU & mDice & mIoU \\
    \midrule
    U-Net\cite{U-Net} & 65.44 & 49.53 & 85.29 & 75.59 \\
    DeepLabV3+\cite{DeepLabV3Plus} & 67.23 & 52.36 & 87.35 & 78.23 \\
    SwinUNet\cite{SwinUNet} & 46.52 & 34.76 & 86.47 & 76.75 \\ 
    H2Former\cite{H2Former} & 67.75 & 52.86 & 86.71 & 77.01 \\
    SAM\cite{SAM} & 25.61 & 14.83 & 27.86 & 16.81 \\
    MedSAM\cite{MedSAM} & 59.89 & 44.06 & 83.67 & 73.63 \\
    SAMed\cite{SAMed} & 49.74 & 34.61 & 80.22 & 68.17 \\
    SAMMed-2D\cite{SAMMed-2D} & 67.22 & 52.47 & 77.49 & 68.07 \\
    MSA\cite{MSA} & 64.16 & 49.50 & 84.09 & 74.34 \\
    SAMUS\cite{SAMUS} & 71.49 & 56.41 & 87.56 & 79.49 \\
    EchoONE & \textbf{73.94} & \textbf{59.31} &  \textbf{88.36} & \textbf{79.65} \\
    \bottomrule
  \end{tabular}
  \caption{Performance of various models on the external datasets HMC\_QU and EchoNet-Dynamic.}
  \label{tab:external_dataset_perform}
\end{table}

\subsubsection{Qualitative analysis}
Examples of segmentation results are visualized in Figure~\ref{fig:all_model_show}. U-Net, DeepLabV3+, SwinUNet, and H2Former are able to segment the general regions, but their performance on the boundaries remains poor. For the SAM-based models, SAM completely loses its segmentation capability on echocardiographic images, while fine-tuned models, including MedSAM, SAMed, SAM-Med2D, MSA, and SAMUS, improve SAM's transferability. However, most of them still exhibit suboptimal segmentation performance. Although SAMUS demonstrated reasonable accuracy in segmenting regions, it still struggled with refining the myocardium contours in multi-plane echocardiography images with significant distribution discrepancies. In contrast, EchoONE generates coarse segmentation masks for different planes to prompt the model to focus on the region and refine the boundaries, achieving improved segmentation results. 

\subsection{Ablation studies}
To further investigate the effectiveness of our dense prompt generation module PC-Mask and the fusion and adaption module (LFFA), ablation studies are conducted on the test data of our five internal datasets. The results are shown in Table \ref{tab:ablation}.

\noindent \textbf{Effectiveness of PC-Mask:} When only PC-Mask is introduced, our method can improve the segmentation performance for all planes, with mIoU gains of 0.06\%, 0.54\%, 1.01\% and 1.48\%, respectively. Especially for PSAX, which differs most from the other three planes, PC-Mask can effectively embed the prior structure through the dense prompt, thereby leading to obvious improvement from 74.86\% to 76.34\% in terms of mIoU. When LFFA is included, PC-Mask can still bring further improvements on all planes. 

\noindent \textbf{Effectiveness of LFFA:} When only LFFA is introduced without PC-Mask,  the performance gains for each plane are 1.11\%, 0.22\%, 0.79\% and 2.48\%, respectively. By integrating local features into the decoder, we have better optimized the model's attention to the boundaries of the cardiac structures, thereby significantly improving the segmentation. Similarly, when PC-Mask is included, LFFA can also bring extra benefits four segmentation of all these planes. 

From these observations, it can be drawn that By leveraging both the prior information in a semantic-aware way and the powerful SAM model with local features fusion and adaption, EchoONE can achieve accurate and robust performance for the MPS problem of echocardiographic images. 



\begin{table}[t]
  \centering
  \begin{tabular}{cccccc}
    \toprule
    \multirow{2}{*}{LFFA} & \multirow{2}{*}{PC-Mask} & \multicolumn{4}{c}{mIoU(\%)} \\ \cline{3-6} 
    \multicolumn{2}{c}{} & 2CH & 3CH & 4CH & PSAX \\
    \midrule
    $\times$ & $\times$ & 80.49 & 77.39 & 81.16 & 74.86 \\
    $\checkmark$ & $\times$ & 81.60 & 77.61 & 81.95 & 77.34 \\
    $\times$ & $\checkmark$ & 80.55 & 77.93 & 82.17 & 76.34 \\
    $\checkmark$ & $\checkmark$ & \textbf{81.87} & \textbf{78.01} & \textbf{82.67} & \textbf{79.68} \\
    \bottomrule
  \end{tabular}
  \caption{Effectiveness of PC-Mask and LFFA in EchoONE.}
  \label{tab:ablation}
\end{table}

%% file: sec/5_consulsion.tex
\section{Conclusion}
In this paper, we propose EchoONE, a model designed to address the challenging multi-plane segmentation problem in echocardiography images. 
We propose a novel dense prompt learning module PC-Mask by leveraging the prior structure knowledge in a composable way, therefore provides effective plane-specific semantic guidance during segmentation. 
We also introduce a learnable CNN local feature branch to tune the image encoder and adapt the mask decoder. A newly-designed LFFA module is introduced for the fusion, which not only enhances the final performance, but also accelerates the convergence. This is the first effective solution that aims to segment all echocardiographic planes with one uniform model. This will definitely simplify the deployment of AI techniques in clinical practice. Although validated only with ultrasound images, our method is also generalizable to MPS problems in other medical image modalities. 


%% file: sec/6_acknowledgement.tex
\section{Acknowledgment}
The work is partially supported by NSFC (62471313, 12326619, 62306183, 62171290), NSF of Guangdong (2024A1515030143, 2024A1515010194, 2023A0505020002), Shenzhen Science and Technology Program (JCYJ20240813141807010), Frontier Technology Development Program of Jiangsu (No.BF2024078), CAMS Innovation Fund for Medical Sciences (CIFMS) (2023-I2M-C\&T-B-117, 2023-I2M-C\&T-B-056), and internal project of National Engineering Laboratory for BDSC (SZU-BDSC-IF2024-02).
